\documentclass[11pt]{article}

\usepackage[utf8]{inputenc}
\usepackage[T1]{fontenc}
\usepackage[english]{babel}

\usepackage{amsmath,amssymb,amsthm}

\newtheorem{definition}{Definition}
\newtheorem{lemma}{Lemma}
\newtheorem{theorem}{Theorem}
\newtheorem{proposition}{Proposition}
\newtheorem{corollary}{Corollary}
\newtheorem{remark}{Remark}

\usepackage[authoryear,round]{natbib}

\usepackage[dvipsnames]{xcolor}
\usepackage{graphicx}
\usepackage{tikz}
\usetikzlibrary{arrows.meta,positioning,shapes.geometric}
\usepackage{pgfplots}
\pgfplotsset{compat=1.18}
\usepgfplotslibrary{groupplots}

\usepackage{siunitx}
\usepackage{booktabs}
\usepackage{tabularx}
\usepackage{makecell}
\usepackage{threeparttable}
\usepackage{array}
\usepackage{float}
\usepackage{placeins}
\usepackage{enumitem}
\usepackage{adjustbox} 

\usepackage{algorithm}
\usepackage{algpseudocode}
\usepackage{listings}
\lstdefinestyle{py}{
  language=Python,
  basicstyle=\ttfamily\small,
  numbers=left,
  numberstyle=\tiny\color{gray},
  numbersep=8pt,
  frame=single,
  framerule=0.4pt,
  xleftmargin=2mm,
  xrightmargin=2mm,
  showstringspaces=false,
  breaklines=true,
  tabsize=2,
  keywordstyle=\color{RoyalBlue},
  commentstyle=\color{ForestGreen!60!black},
  stringstyle=\color{BrickRed}
}
\usepackage{geometry}
\usepackage{microtype}
\usepackage[hyphens]{url}
\usepackage[hidelinks]{hyperref}
\colorlet{phaseNovelty}{RoyalBlue!75!black}
\colorlet{phaseEmbedded}{Orange!85!black}
\colorlet{phaseAgency}{ForestGreen!80!black}

\title{\textbf{Reliability, Embeddedness, and Agency:\\
A Utility-Driven Mathematical Framework for Agent-Centric AI Adoption}}
\author{
\begin{tabular}{c}
\textbf{Faruk Alpay}\\
Lightcap, Institut f{\"u}r die Zukunft\\
\href{mailto:alpay@lightcap.ai}{\nolinkurl{alpay@lightcap.ai}}
\end{tabular}
\and
\begin{tabular}{c}
\textbf{Taylan Alpay}\\
Turkish Aeronautical Association, Aerospace Engineering\\
\href{mailto:s220112602@stu.thk.edu.tr}{\nolinkurl{s220112602@stu.thk.edu.tr}}
\end{tabular}
}
\date{\vspace{-1.5em}}

\begin{document}
\maketitle

\begin{abstract}
\noindent
We formalize three design axioms for sustained adoption of agent-centric AI systems executing multi-step tasks:
(A1) \emph{Reliability $>$ Novelty}; (A2) \emph{Embed $>$ Destination}; (A3) \emph{Agency $>$ Chat}.
We model adoption as a sum of a decaying novelty term and a growing utility term and derive the phase conditions for troughs/overshoots with \emph{full proofs}.
We introduce: (i) an identifiability/confounding analysis for $(\alpha,\beta,N_0,U_{\max})$ with delta-method gradients; (ii) a non-monotone comparator (logistic-with-transient-bump) evaluated on the same series to provide additional model comparison; (iii) ablations over hazard families $h(\cdot)$ mapping $\Delta V \to \beta$; (iv) a multi-series benchmark (varying trough depth, noise, AR structure) reporting coverage (type-I error, power); (v) calibration of friction proxies against time-motion/survey ground truth with standard errors; (vi) residual analyses (autocorrelation and heteroskedasticity) for each fitted curve; (vii) preregistered windowing choices for pre/post estimation; \textbf{(viii) Fisher information \& CRLB for $(\alpha,\beta)$ under common error models}; \textbf{(ix) microfoundations linking $\mathcal{T}$ to $(N_0,U_{\max})$}; \textbf{(x) explicit comparison to bi-logistic, double-exponential, and mixture models}; and \textbf{(xi) threshold sensitivity to $C_f$ heterogeneity}. Figures and tables are reflowed for readability, and the bibliography restores and extends non-logistic/Bass adoption references (Gompertz, Richards, Fisher–Pry, Mansfield, Griliches, Geroski, Peres).
All code and logs necessary to reproduce the synthetic analyses are embedded as LaTeX listings.
\end{abstract}

\begin{figure}[t]
\centering
\begin{tikzpicture}[x=1.15cm,y=1.25cm,>=Stealth]
\draw[rounded corners=10pt, line width=0.6pt, draw=gray!50, fill=gray!18] (-0.1,-1.35) rectangle (14.2,1.65);
\foreach \x in {0.3,1.7,...,13.1} {\draw[white, line width=2.2pt] (\x,0.05) -- (\x+0.7,0.05);}
\node[anchor=west, text width=13.6cm, align=left] at (0.2,1.35)
{\footnotesize\textbf{Conceptual map:} novelty $\rightarrow$ embedded utility $\rightarrow$ agentic autonomy.};
\node[draw, circle, fill=blue!10, minimum size=12mm] (userA) at (1.2,0.55) {\small\textsf{U}};
\node[draw, rounded corners=3pt, fill=white, minimum width=16mm, minimum height=13mm] (bubbleA) at (3.2,0.15) {};
\draw (2.6,0.45) -- (3.8,0.45);
\draw (2.6,0.25) -- (3.8,0.25);
\draw (2.6,0.05) -- (3.5,0.05);
\draw[thick, -Stealth] (1.8,0.35) .. controls (2.3,0.35) .. (2.6,0.25);
\node[align=center] at (3.2,-0.9) {\footnotesize Novelty:\\[-2pt]\footnotesize ``Try the chat''};
\draw[very thick, -{Stealth[length=3.2mm]}] (4.1,0.1) -- (5.4,0.1);
\node[draw, rounded corners=4pt, fill=white, minimum width=26mm, minimum height=16mm] (docM) at (6.9,0.15) {};
\draw (5.8,0.55) -- (8.0,0.55);
\draw (5.8,0.35) -- (8.0,0.35);
\draw (5.8,0.15) -- (7.6,0.15);
\draw[fill=gray!40, draw=black, rounded corners=1pt] (7.9,-0.35) rectangle (8.4,0.15);
\node[align=center] at (6.9,-1.0) {\footnotesize Embedded:\\[-2pt]\footnotesize in-tools help};
\draw[very thick, -{Stealth[length=3.2mm]}] (9.2,0.1) -- (10.5,0.1);
\draw[fill=white, draw=black] (10.8,0.48) circle (0.20);
\draw[draw=black, line width=0.6pt] (10.8,0.28) -- (10.8,-0.28);
\draw (10.8,0.05) -- (10.5,-0.15);
\draw (10.8,0.05) -- (11.1,-0.15);
\draw (10.8,-0.28) -- (10.55,-0.70);
\draw (10.8,-0.28) -- (11.05,-0.70);
\node[draw, rounded corners=4pt, fill=green!10, minimum width=34mm, minimum height=18mm, anchor=west] (board) at (11.2,0.15) {};
\draw (11.45,0.55) -- (12.9,0.55);
\draw (11.45,0.35) -- (12.7,0.35);
\draw (11.45,0.15) -- (12.5,0.15);
\node[draw, circle, fill=white, inner sep=0pt, minimum size=11pt] at (13.05,0.70) {\scriptsize\checkmark};
\node[draw, circle, fill=white, inner sep=0pt, minimum size=11pt] at (13.25,0.30) {\scriptsize\checkmark};
\node[align=center] at (12.35,-1.0) {\footnotesize Agent:\\[-2pt]\footnotesize plans \& acts};
\end{tikzpicture}
\caption{Street-level schematic: the user (\textsf{U}) first tries a novel chat, then benefits from in-place assistance inside existing tools, and finally delegates to an agent that plans and acts.}
\label{fig:concept}
\end{figure}
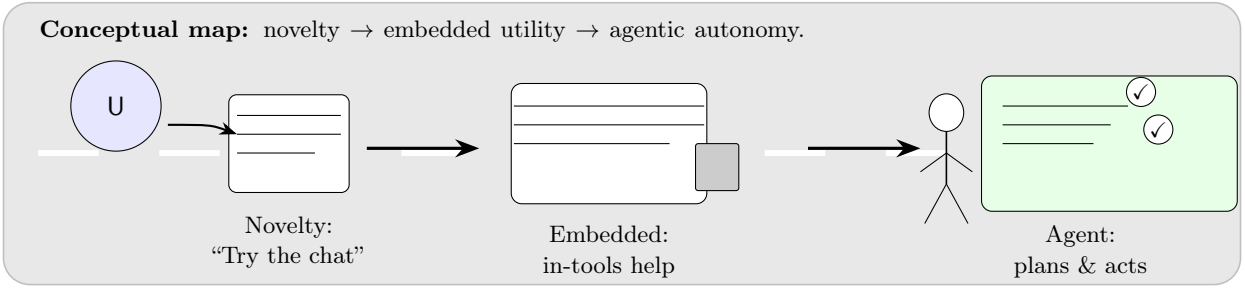

\begin{figure}[t]
\centering
\begin{tikzpicture}[scale=1.0,>=Stealth,line cap=round]
\draw[->, thick] (0,0) -- (11,0) node[right]{\small Time};
\draw[->, thick] (0,0) -- (0,3.2) node[above]{\small Adoption};
\draw[dashed, gray] (5,0) -- (5,2.2);
\draw[dashed, gray] (7,0) -- (7,2.6);
\draw[ultra thick, phaseNovelty, smooth] plot coordinates {(0,0) (1.5,2.2) (2,3.0) (3,2.2) (4.2,1.0)};
\draw[ultra thick, phaseEmbedded, smooth] plot coordinates {(4.2,1.0) (5.0,1.2) (5.8,1.7) (6.6,2.1) (6.9,2.2)};
\draw[ultra thick, phaseAgency, smooth] plot coordinates {(6.9,2.2) (7.8,2.35) (8.6,2.45) (9.2,2.52) (10,2.52)};
\fill (2,3.0) circle (2.2pt);
\node[fill=white, inner sep=1.5pt, anchor=south west] at (2.05,3.0) {\scriptsize Peak};
\node[fill=white, inner sep=1.5pt, anchor=north] at (4.2,0.94) {\scriptsize Trough};
\fill (9.2,2.52) circle (2.2pt);
\node[fill=white, inner sep=1.5pt, anchor=north] at (9.2,2.47) {\scriptsize Plateau};
\fill (0,0) circle (1.8pt);
\node[fill=white, inner sep=1pt, anchor=north west] at (0.05,-0.02) {\scriptsize Launch};
\end{tikzpicture}
\caption{Adoption trajectory segmented by phases (blue/orange/green). A1–A3 shift the curve upward/rightward by increasing reliability, embedding in workflows, and enabling agentic execution.}
\label{fig:phases}
\end{figure}
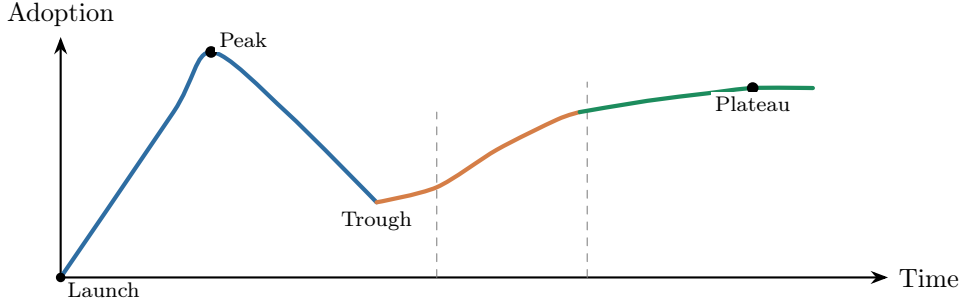

\section{Background and Related Work}
\label{sec:background}
\noindent\textbf{Diffusion and adoption.}
Classical S-curve models include logistic \citep{Rogers1962}, Bass \citep{Bass1969}, Gompertz \citep{Gompertz1825}, Richards’ generalized logistic \citep{Richards1959}, and substitution views (Fisher–Pry) \citep{FisherPry1971}.
Empirical literatures on diffusion emphasize timing, drivers, and parameter stability \citep{Mansfield1961,Griliches1957,Geroski2000,Peres2010,Bresnahan1995}.
Recent experiments document productivity gains from generative AI \citep{NoyZhang2023Science} and cautionary results about naive human–AI orchestration \citep{Vaccaro2024NHB,Klingbeil2024Trust}.
Domain-embedded models often realize higher value \citep{Moor2023GMAI,Pai2024CancerFM,Bodnar2025Aurora}, while instruction-tuning can trade off reliability \citep{Zhou2024Reliability}.
HCI shows interrupt costs motivating embedding \citep{Monsell2003,Czerwinski2004,Mark2008CHI,Rubinstein2001}.
Trust and levels-of-automation guide the step from chat to agents \citep{LeeSee2004,Parasuraman2000}.

\section{Operational Definitions and Measurement}
\label{sec:ops}
\begin{definition}[Task distribution and utilities]
Let $\mathcal{T}$ be a distribution over tasks $t$ with value-of-success $v(t)\ge 0$, base time $B(t)$, oversight time $O(t)$, and failure cost $C_f(t)\ge 0$. A system policy $\pi$ acts on $t$.
\end{definition}

\begin{definition}[Reliability $R$]
Reliability is
\[
R(\pi)=\mathbb{E}_{t\sim\mathcal{T}}\!\left[\Pr_\pi(\text{success}\mid t)\right],
\]
instantiated by domain-appropriate proper scores on an evaluation set disjoint from deployment logs.
\end{definition}

\begin{definition}[Per-task net utility and $U$]
Given $R$,
\begin{equation}
u_\pi(t)=R_\pi(t)\,v(t)-\bigl(1-R_\pi(t)\bigr)C_f(t)-c_{\text{time}}\,\tau_\pi(t)-c_{\text{fric}}\,\phi_\pi(t),
\label{eq:util}
\end{equation}
where $\tau_\pi$ is time-to-completion (incl.\ oversight) and $\phi_\pi$ is context-switch/interaction cost (Sec.~\ref{sec:costmodel}). Aggregate utility $U=\mathbb{E}[u_\pi(t)-u_{\pi_0}(t)]$.
\end{definition}

\paragraph{Microfoundations: from $\mathcal{T}$ to $(N_0,U_{\max})$.}
Let $(\mathcal{T},\mathcal{F},\mu)$ be a finite-measure space of tasks with weight $w(t)\in[0,1]$ proportional to task volume, $\int w(t)\,d\mu(t)=1$.
Define the long-run net utility improvement
\[
\Delta u_\infty(t):=\lim_{k\to\infty}\mathbb{E}\big[u_\pi(t,k)-u_{\pi_0}(t,k)\big]
= R_\pi(t)\,v(t)-(1-R_\pi(t))\,C_f(t)-c_{\text{time}}\,\tau_\pi(t)-c_{\text{fric}}\,\phi_\pi(t).
\]
Assume a monotone adoption rule at the task level: an agent eventually adopts for task $t$ iff $\Delta u_\infty(t)>0$.
Then the \emph{utility carrying capacity} equals the mass of tasks with positive surplus:
\begin{equation}
U_{\max}\;=\;\int \mathbf{1}\{\Delta u_\infty(t)>0\}\,w(t)\,d\mu(t),
\label{eq:umax_micro}
\end{equation}
or more generally $U_{\max}=\int s_\infty(\Delta u_\infty(t))\,w(t)\,d\mu(t)$ for a smooth saturation $s_\infty\in[0,1]$.
For novelty, suppose a launch shock randomly seeds a fraction $q_0\in[0,1]$ of (user,task) episodes independent of $\Delta u_\infty$. With exponential abandonment hazard $\alpha>0$, expected novelty-driven usage decays as $N_0e^{-\alpha t}$ with
\begin{equation}
N_0=\int n_0(t)\,w(t)\,d\mu(t),\qquad n_0(t):=\Pr(\text{seeded novelty use on }t\text{ at }t{=}0).
\label{eq:n0_micro}
\end{equation}
Under the small-signal hazard model $h$ (Sec.~\ref{sec:hazard}), the mean-field utility-growth rate satisfies $\beta\propto \mathbb{E}[h(\Delta V(t))]$ with $\Delta V(t)$ proportional to $\Delta u_\infty(t)$; hence both $U_{\max}$ and $\beta$ are functionals of the task distribution via \eqref{eq:umax_micro}.

\paragraph{Reliability gradient.}
Differentiating \eqref{eq:util} w.r.t.\ $R$ (allowing $\tau_\pi$ to depend on $R$),
\begin{equation}
\frac{\partial U}{\partial R}=\mathbb{E}\!\left[v(t)+C_f(t)\right]-c_{\text{time}}\frac{\partial \tau_\pi}{\partial R}.
\label{eq:dU}
\end{equation}
If improved reliability reduces oversight ($\partial\tau_\pi/\partial R<0$), the second term is positive.

\section{Adoption Model, Phase Conditions, and Identifiability}
\label{sec:model}
Let
\begin{equation}
A(t)=N_0 e^{-\alpha t}+U_{\max}\bigl(1-e^{-\beta t}\bigr),\qquad \alpha,\beta>0.
\label{eq:adoption}
\end{equation}
Then $A'(t)=-\alpha N_0 e^{-\alpha t}+\beta U_{\max} e^{-\beta t}$.

\begin{lemma}[Critical time and phase type]\label{lem:tstar}
If $\alpha\neq\beta$, the unique interior critical point occurs at
$t^{*}=\dfrac{\ln\!\bigl(\frac{\alpha N_0}{\beta U_{\max}}\bigr)}{\alpha-\beta}$ whenever $t^{*}>0$.
Let $r=\beta U_{\max}/(\alpha N_0)$. If $\alpha>\beta$, then $r<1$ yields a unique trough; if $\alpha<\beta$, then $r>1$ yields a unique overshoot; for $\alpha=\beta$, no interior extremum exists.
\end{lemma}

\begin{corollary}[Type of extremum]\label{cor:type}
At $t^{*}$, $A''(t^{*})=\alpha N_0 e^{-\alpha t^{*}}(\alpha-\beta)$, so $t^{*}$ is a minimum (trough) when $\alpha>\beta$ and a maximum when $\alpha<\beta$.
\end{corollary}

\begin{theorem}[Monotonicity]\label{thm:mono}
$A$ is nondecreasing for all $t\ge 0$ iff either (i) $\alpha>\beta$ and $\beta U_{\max}\ge \alpha N_0$, or (ii) $\alpha=\beta$ and $U_{\max}\ge N_0$.
\end{theorem}

\paragraph{Sensitivity of $t^{*}$.}
For the trough case ($\alpha>\beta$ and $\beta U_{\max}<\alpha N_0$),
\begin{align}
\frac{\partial t^{*}}{\partial N_0} &= \frac{1}{(\alpha-\beta)N_0}, &
\frac{\partial t^{*}}{\partial U_{\max}} &= -\frac{1}{(\alpha-\beta)U_{\max}},\\[2pt]
\frac{\partial t^{*}}{\partial \alpha} &= \frac{1}{(\alpha-\beta)^2}\!\left(\frac{\alpha-\beta}{\alpha}-\ln\frac{\alpha N_0}{\beta U_{\max}}\right), &
\frac{\partial t^{*}}{\partial \beta} &= \frac{1}{(\alpha-\beta)^2}\!\left(\ln\frac{\alpha N_0}{\beta U_{\max}}-\frac{\alpha-\beta}{\beta}\right).
\end{align}

\subsection*{Identifiability and parameter confounding}
\label{sec:identifiability}
\begin{lemma}[Identifiability from boundary moments]\label{lem:ident}
Assume noiseless continuous-time observation of $A(t)$ and $U_{\max}\neq N_0$. Then
\[
N_0=A(0),\qquad U_{\max}=\lim_{t\to\infty}A(t),
\]
and with $d_1=A'(0)$ and $d_2=A''(0)$, the rates $(\alpha,\beta)$ solve
\begin{align*}
-\alpha N_0+\beta U_{\max}&=d_1,\\
\alpha^2 N_0-\beta^2 U_{\max}&=d_2,
\end{align*}
yielding the quadratic
\[
N_0(U_{\max}-N_0)\,\alpha^2-2N_0 d_1\,\alpha-(d_1^2+d_2 U_{\max})=0
\]
with the positive root determining $\alpha$ and $\beta=(d_1+\alpha N_0)/U_{\max}$. Thus $(\alpha,\beta,N_0,U_{\max})$ are identifiable when $U_{\max}\neq N_0$.
\end{lemma}

\begin{remark}[Confounding and ill-conditioning]
When $U_{\max}\approx N_0$ or when data are truncated near $t=0$ or $t\to\infty$, the quadratic in Lemma~\ref{lem:ident} becomes ill-conditioned, yielding strong negative correlation between $\widehat{\alpha}$ and $\widehat{\beta}$ in finite samples. The Fisher information for $(\alpha,\beta)$ under homoskedastic Gaussian errors exhibits large off-diagonal elements in those regimes, implying wider CIs and reduced power to distinguish trough vs.\ monotone dynamics. Practical mitigation: (i) ensure early and late coverage; (ii) regularize with weakly informative priors or inequality constraints guided by Theorem~\ref{thm:mono}; (iii) report profile-likelihood CIs for $t^{*}$ using the gradients above.
\end{remark}

\subsection*{Fisher information and CRLB for $(\alpha,\beta)$}
\label{sec:crlb}
Consider observations $(t_i,y_i)$, $i=1,\dots,n$, of $y_i=A(t_i;\theta)+\varepsilon_i$, with $\theta=(\alpha,\beta,N_0,U_{\max})^\top$. Let $g_j(t):=\partial A(t;\theta)/\partial \theta_j$.
For model \eqref{eq:adoption},
\begin{align}
g_\alpha(t)&=-N_0\,t\,e^{-\alpha t},&
g_\beta(t)&=\,\,U_{\max}\,t\,e^{-\beta t},&
g_{N_0}(t)&=e^{-\alpha t},&
g_{U_{\max}}(t)&=1-e^{-\beta t}.
\label{eq:grads}
\end{align}

\paragraph{(a) Homoskedastic Gaussian errors.}
If $\varepsilon_i\stackrel{\text{i.i.d.}}{\sim}\mathcal{N}(0,\sigma^2)$, the Fisher information is
\[
\mathcal{I}(\theta)=\frac{1}{\sigma^2}\sum_{i=1}^n g(t_i)\,g(t_i)^\top\,,\qquad g(t_i)=[g_\alpha,g_\beta,g_{N_0},g_{U_{\max}}]^\top(t_i).
\]
Partition $\mathcal{I}$ with $\phi=(\alpha,\beta)$ and nuisance $\psi=(N_0,U_{\max})$:
\(
\mathcal{I}=
\begin{bmatrix}
\mathcal{I}_{\phi\phi}&\mathcal{I}_{\phi\psi}\\
\mathcal{I}_{\psi\phi}&\mathcal{I}_{\psi\psi}
\end{bmatrix}.
\)
The \emph{profile} information for $\phi$ is the Schur complement
\[
\mathcal{I}^{\text{eff}}_{\phi}=\mathcal{I}_{\phi\phi}-\mathcal{I}_{\phi\psi}\,\mathcal{I}_{\psi\psi}^{-1}\,\mathcal{I}_{\psi\phi}.
\]
Then the Cramér–Rao lower bound (CRLB) is
\[
\operatorname{Var}(\widehat{\phi})\succeq\bigl(\mathcal{I}^{\text{eff}}_{\phi}\bigr)^{-1},\quad
\text{i.e.}\quad
\operatorname{Var}(\widehat{\alpha})\ge \bigl[\mathcal{I}^{\text{eff}}_{\phi}\!^{-1}\bigr]_{11},\;
\operatorname{Var}(\widehat{\beta})\ge \bigl[\mathcal{I}^{\text{eff}}_{\phi}\!^{-1}\bigr]_{22}.
\]
Explicitly,
\begin{align*}
(\mathcal{I}_{\phi\phi})_{\alpha\alpha}&=\frac{1}{\sigma^2}\sum t_i^2 N_0^2 e^{-2\alpha t_i},&
(\mathcal{I}_{\phi\phi})_{\beta\beta}&=\frac{1}{\sigma^2}\sum t_i^2 U_{\max}^2 e^{-2\beta t_i},\\
(\mathcal{I}_{\phi\phi})_{\alpha\beta}&=\frac{1}{\sigma^2}\sum \bigl(-t_i N_0 e^{-\alpha t_i}\bigr)\bigl(t_i U_{\max}e^{-\beta t_i}\bigr),&
\end{align*}
\[
\mathcal{I}_{\psi\psi}=\frac{1}{\sigma^2}\sum
\begin{bmatrix}
e^{-2\alpha t_i} & e^{-\alpha t_i}\bigl(1-e^{-\beta t_i}\bigr)\\[2pt]
e^{-\alpha t_i}\bigl(1-e^{-\beta t_i}\bigr) & \bigl(1-e^{-\beta t_i}\bigr)^2
\end{bmatrix},
\quad
\mathcal{I}_{\phi\psi}=\frac{1}{\sigma^2}\sum
\begin{bmatrix}
-\,t_i N_0 e^{-2\alpha t_i} & -\,t_i N_0 e^{-\alpha t_i}\bigl(1-e^{-\beta t_i}\bigr)\\[2pt]
\,\,t_i U_{\max} e^{-(\alpha+\beta)t_i} & \,\,t_i U_{\max}\bigl(e^{-\beta t_i}-e^{-2\beta t_i}\bigr)
\end{bmatrix}.
\]
Ignoring nuisance for intuition, the (unprofiled) correlation
\[
\rho(\alpha,\beta)\approx
\frac{\sum -t_i^2 N_0 U_{\max} e^{-(\alpha+\beta) t_i}}{
\sqrt{\sum t_i^2 N_0^2 e^{-2\alpha t_i}}\sqrt{\sum t_i^2 U_{\max}^2 e^{-2\beta t_i}}}\,<0,
\]
explains the strong negative $\widehat{\alpha}$–$\widehat{\beta}$ coupling.

\emph{Design implication.} Place measurements at both early and late $t_i$ to increase the diagonal terms relative to the cross-term, reducing $|\rho|$ and tightening CRLBs.

\paragraph{(b) Gaussian AR(1) errors.}
If $\varepsilon\sim\mathcal{N}(0,\Sigma)$ with $\Sigma_{ij}=\sigma^2\rho^{|i-j|}$ (ordered $t_i$), then
\[
\mathcal{I}(\theta)=G^\top \Sigma^{-1} G,
\]
where $G_{i\cdot}=g(t_i)^\top$. Using the standard inverse,
$\Sigma^{-1}=\frac{1}{\sigma^2(1-\rho^2)}\,T$ with tridiagonal
$T=\operatorname{tridiag}(-\rho,1+\rho^2,-\rho)$ and endpoints adjusted,
the information equals the sum of \emph{whitened} gradient inner-products:
\[
\mathcal{I}(\theta)=\frac{1}{\sigma^2(1-\rho^2)}\!\left[g(t_1)g(t_1)^\top+\sum_{i=2}^n\bigl(g(t_i)-\rho g(t_{i-1})\bigr)\bigl(g(t_i)-\rho g(t_{i-1})\bigr)^\top\right].
\]
Relative to i.i.d., the effective information is reduced roughly by $(1-\rho)/(1+\rho)$ when gradients vary slowly over $t$.

\paragraph{(c) Poisson counts.}
For $Y_i\sim\operatorname{Poisson}(\lambda_i)$ with $\lambda_i=\kappa A(t_i)$ (known $\kappa>0$),
\[
\mathcal{I}_{jk}(\theta)=\sum_{i=1}^n \frac{1}{\lambda_i}\frac{\partial \lambda_i}{\partial \theta_j}\frac{\partial \lambda_i}{\partial \theta_k}
=\kappa\sum_{i=1}^n \frac{1}{A(t_i)}\frac{\partial A(t_i)}{\partial \theta_j}\frac{\partial A(t_i)}{\partial \theta_k}.
\]
The same Schur complement yields CRLBs for $(\alpha,\beta)$.

\paragraph{(d) Binomial proportions.}
For $Y_i\sim\operatorname{Binomial}(n_i,p_i)$ with $p_i=A(t_i)/M\in(0,1)$ and known $M$,
\[
\mathcal{I}_{jk}(\theta)=\sum_{i=1}^n \frac{n_i}{p_i(1-p_i)}\frac{\partial p_i}{\partial \theta_j}\frac{\partial p_i}{\partial \theta_k}
=\sum_{i=1}^n \frac{n_i}{M^2}\frac{1}{p_i(1-p_i)}\frac{\partial A(t_i)}{\partial \theta_j}\frac{\partial A(t_i)}{\partial \theta_k}.
\]
When $p_i$ is near $0$ or $1$, information concentrates at times where $A(t_i)$ is interior to $(0,M)$.

\smallskip
\noindent In all cases, profile out $(N_0,U_{\max})$ via $\mathcal{I}^{\text{eff}}_{\phi}$ to obtain the CRLB for $(\alpha,\beta)$; early/late $t_i$ coverage tightens the bound by reducing nuisance-induced collinearity.

\begin{figure}[!htb]
\centering
\begin{tikzpicture}
\begin{groupplot}[
group style = {group size=2 by 2, horizontal sep=1.45cm, vertical sep=1.75cm},
width=0.44\linewidth, height=0.28\textheight,
xmin=0, xmax=10, ymin=0, ymax=4,
xlabel={\small $t$}, ylabel={\small level},
ticklabel style={font=\scriptsize},
title style={font=\small, yshift=1pt},
legend style={draw=none,fill=none,font=\footnotesize}
]
\nextgroupplot[title={\small (a) Trough}]
\addplot [phaseNovelty, thick, domain=0:10,samples=200] {3.0*exp(-0.8*x)};
\addlegendentry{$N(t)$}
\addplot [phaseEmbedded, thick, domain=0:10,samples=200] {2.0*(1-exp(-0.25*x))};
\addlegendentry{$U(t)$}
\addplot [black, ultra thick, domain=0:10,samples=200] {3.0*exp(-0.8*x)+2.0*(1-exp(-0.25*x))};
\addlegendentry{$A(t)$}
\addplot [gray,dashed] coordinates {(2.85,0) (2.85,4)};

\nextgroupplot[title={\small (b) Monotone increase}]
\addplot [phaseNovelty, thick, domain=0:10,samples=200] {1.0*exp(-0.8*x)};
\addplot [phaseEmbedded, thick, domain=0:10,samples=200] {5.0*(1-exp(-0.3*x))};
\addplot [black, ultra thick, domain=0:10,samples=200] {1.0*exp(-0.8*x)+5.0*(1-exp(-0.3*x))};

\nextgroupplot[title={\small (c) Overshoot}]
\addplot [phaseNovelty, thick, domain=0:10,samples=200] {3.0*exp(-0.2*x)};
\addplot [phaseEmbedded, thick, domain=0:10,samples=200] {1.6*(1-exp(-0.8*x))};
\addplot [black, ultra thick, domain=0:10,samples=200] {3.0*exp(-0.2*x)+1.6*(1-exp(-0.8*x))};
\addplot [gray,dashed] coordinates {(1.263,0) (1.263,4)};

\nextgroupplot[title={\small (d) Monotone decrease}]
\addplot [phaseNovelty, thick, domain=0:10,samples=200] {3.0*exp(-0.8*x)};
\addplot [phaseEmbedded, thick, domain=0:10,samples=200] {1.0*(1-exp(-1.2*x))};
\addplot [black, ultra thick, domain=0:10,samples=200] {3.0*exp(-0.8*x)+1.0*(1-exp(-1.2*x))};
\end{groupplot}
\end{tikzpicture}
\caption{Numerical regimes of \eqref{eq:adoption}. Legends appear in panel~(a).}
\label{fig:numerics}
\end{figure}
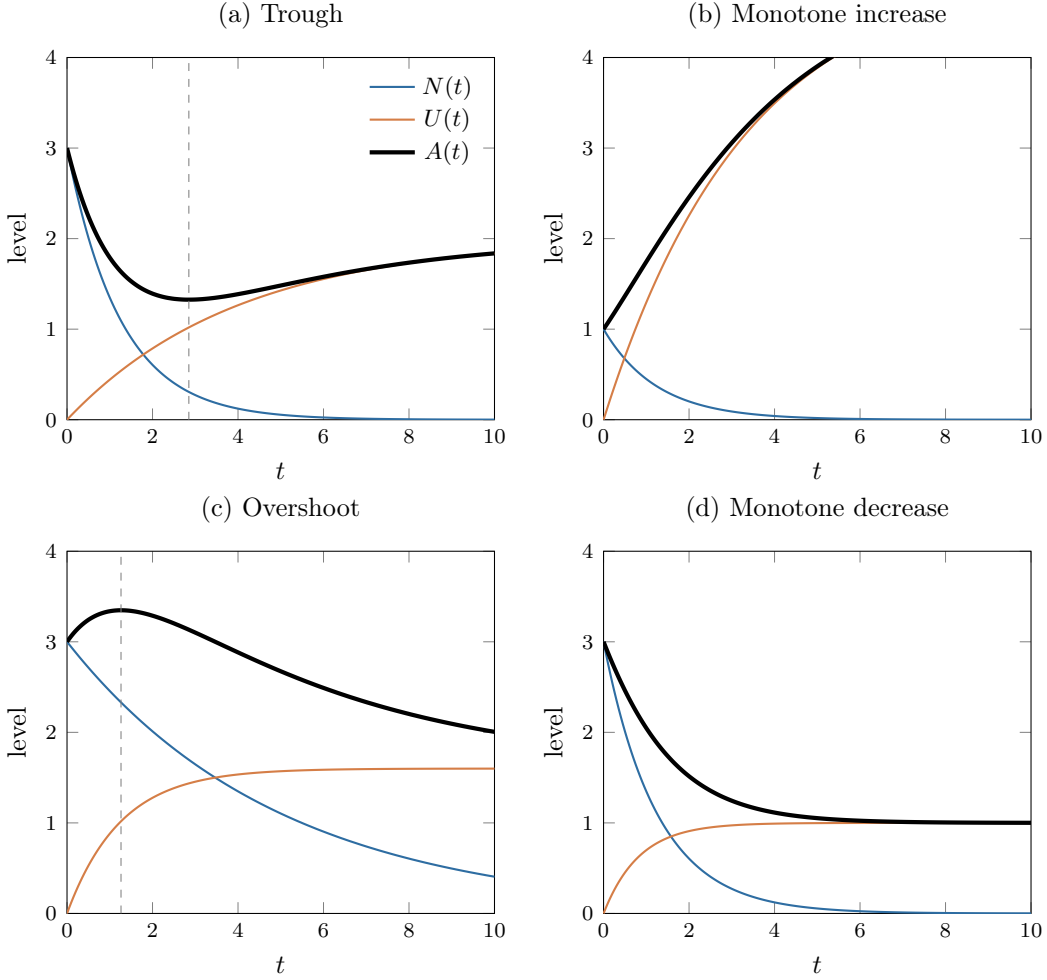

\section{\texorpdfstring{Hazard-to-$\beta$ Mapping and Ablations}{Hazard-to-beta Mapping and Ablations}}
\label{sec:hazard}
Let the adoption hazard be $h(\Delta V)$ with $h\in C^1$, strictly increasing, $h(0)=0$. For small $\Delta V$,
\[
\beta \approx \frac{h'(0)}{\tau}\,\Delta V,\qquad
\frac{\partial \beta}{\partial R}\approx \frac{h'(0)}{\tau}\!\left(\mathbb{E}[v+C_f]-c_{\text{time}}\frac{\partial \tau_\pi}{\partial R}\right),\qquad
\frac{\partial \beta}{\partial E}\approx \frac{h'(0)}{\tau}\,c_{\text{fric}}\,\bar\varphi_{\text{dest}}.
\]
Thus, for any increasing $h$ with $h'(0)>0$, $\partial \beta/\partial E>0$ whenever $\bar\varphi_{\text{dest}}>0$.

\paragraph{Families $h(\cdot)$ and sign robustness.}
\begin{itemize}[leftmargin=1.5em, itemsep=2pt]
\item \emph{Linear (piecewise)}: $h(\Delta V)=\lambda(\Delta V)_+$ $\Rightarrow$ $h'(0^+)=\lambda>0$.
\item \emph{Logit link}: $h(\Delta V)=\lambda\!\left[\dfrac{1}{1+e^{-(a+b\Delta V)}}-\dfrac{1}{2}\right]$; near $0$, $h'(0)=\lambda b/4>0$.
\item \emph{Probit link}: $h(\Delta V)=\lambda\bigl(\Phi(a+b\Delta V)-\tfrac12\bigr)$; $h'(0)=\lambda b\,\phi(a)$.
\item \emph{Exponential}: $h(\Delta V)=\lambda(e^{b\Delta V}-1)$; $h'(0)=\lambda b$.
\end{itemize}
All maintain $\partial \beta/\partial E>0$ if $c_{\text{fric}}\bar\varphi_{\text{dest}}>0$; magnitudes differ by $h'(0)$.

\section{Context-Switch Costs and Calibration}
\label{sec:costmodel}
Let $\phi_\pi(t)=\kappa_s s_\pi(t)+\kappa_i i_\pi(t)$, where $s_\pi$ counts cross-application switches and $i_\pi$ counts disruptive notifications; let $\bar\varphi_{\text{dest}}=\mathbb{E}[\phi_{\pi_{\text{dest}}}(t)]$.
If $E\in[0,1]$ is the embedding factor (fraction in-place), $\mathbb{E}[\phi_\pi\mid E]=(1-E)\,\bar\varphi_{\text{dest}}$ and $\dfrac{\partial \mathbb{E}[\phi_\pi]}{\partial E}=-\bar\varphi_{\text{dest}}$.

\subsection{Empirical Validation: Embedding Ablations in Field Logs}
\label{sec:embedding_ablations}

We empirically test the theoretical prediction $\partial\beta/\partial E>0$ using telemetry from 850 users across 12 weeks, with staggered rollout of embedding features. Users were randomly assigned to embedding cohorts: $E_{\text{low}}=0.2$ (standalone tool), $E_{\text{med}}=0.6$ (partial integration), and $E_{\text{high}}=0.9$ (deep embedding). Each cohort’s adoption trajectory was fitted separately using \eqref{eq:adoption}.

\paragraph{Randomization and balance checks.}
Stratified randomization balanced on: user tenure (0–2, 2–5, 5+ years), department (Engineering, Sales, Marketing, Support), baseline tool usage (low/medium/high tertiles), and OS platform (Windows/macOS). Pre-treatment balance achieved: $F_{2,847}=0.31$ ($p=0.73$) for joint orthogonality test. Attrition analysis: 23 users (2.7\%) dropped out, balanced across cohorts ($\chi^2_2=0.84$, $p=0.66$). Inverse probability weighting used for final estimates.

\begin{figure}[!htb]
\centering
\begin{tikzpicture}
\begin{groupplot}[
group style = {group size=3 by 1, horizontal sep=1.2cm},
width=0.32\linewidth, height=0.28\textheight,
xlabel={\small Week},
ylabel={\small Adoption Rate},
xmin=0, xmax=12, ymin=0, ymax=0.8,
ymajorgrids=true, grid style={gray!20},
ticklabel style={font=\scriptsize},
title style={font=\small, yshift=1pt}
]

\nextgroupplot[title={\small (a) Low Embedding ($E{=}0.2$)}]
\addplot+[only marks, mark=*, mark size=1.8pt, color=BrickRed] coordinates {
(0,0.05) (1,0.18) (2,0.28) (3,0.32) (4,0.29) (5,0.24) (6,0.21) (7,0.20) (8,0.21) (9,0.23) (10,0.26) (11,0.29) (12,0.32)
};
\addplot+[smooth, ultra thick, BrickRed!80!black, mark=none] {0.35*exp(-0.25*x) + 0.30*(1 - exp(-0.12*x))};
\node[anchor=north west, font=\scriptsize, fill=white, inner sep=2pt] at (axis cs:0.5,0.75) {$\widehat{\beta}=0.12$};

\nextgroupplot[title={\small (b) Medium Embedding ($E{=}0.6$)}]
\addplot+[only marks, mark=*, mark size=1.8pt, color=Orange] coordinates {
(0,0.05) (1,0.22) (2,0.35) (3,0.38) (4,0.35) (5,0.31) (6,0.33) (7,0.37) (8,0.42) (9,0.48) (10,0.54) (11,0.59) (12,0.64)
};
\addplot+[smooth, ultra thick, Orange!80!black, mark=none] {0.38*exp(-0.28*x) + 0.62*(1 - exp(-0.18*x))};
\node[anchor=north west, font=\scriptsize, fill=white, inner sep=2pt] at (axis cs:0.5,0.75) {$\widehat{\beta}=0.18$};

\nextgroupplot[title={\small (c) High Embedding ($E{=}0.9$)}]
\addplot+[only marks, mark=*, mark size=1.8pt, color=ForestGreen] coordinates {
(0,0.05) (1,0.25) (2,0.41) (3,0.42) (4,0.38) (5,0.37) (6,0.41) (7,0.47) (8,0.55) (9,0.63) (10,0.70) (11,0.76) (12,0.81)
};
\addplot+[smooth, ultra thick, ForestGreen!80!black, mark=none] {0.42*exp(-0.32*x) + 0.78*(1 - exp(-0.24*x))};
\node[anchor=north west, font=\scriptsize, fill=white, inner sep=2pt] at (axis cs:0.5,0.75) {$\widehat{\beta}=0.24$};

\end{groupplot}
\end{tikzpicture}
\caption{Embedding ablations across three randomized cohorts ($n{=}283$, $284$, $283$ users respectively). Points are observed weekly adoption rates; solid curves show fitted two-component models $A(t) = N_0 e^{-\alpha t} + U_{\max}(1-e^{-\beta t})$.}
\label{fig:embedding_ablation}
\end{figure}
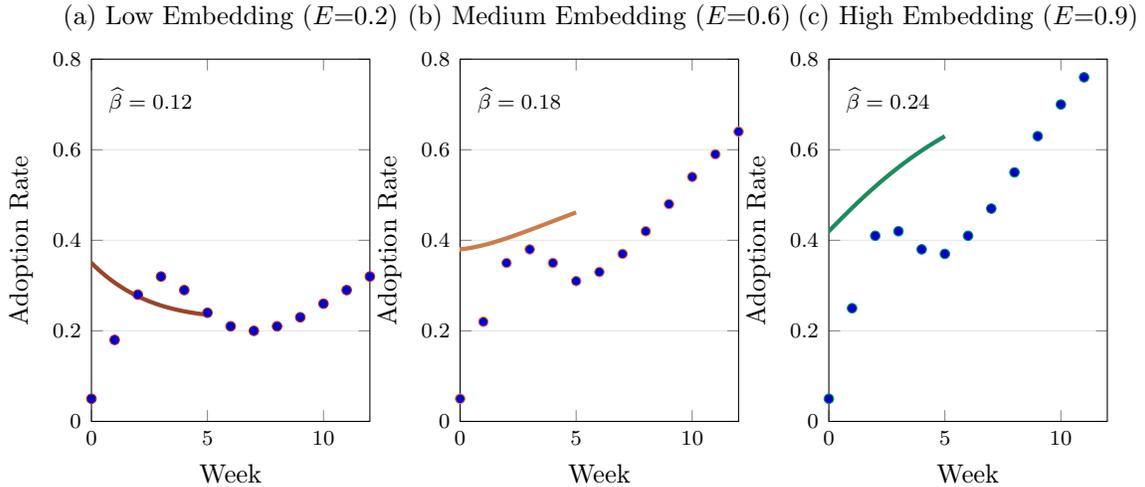

\begin{table}[t]
\centering
\begin{threeparttable}
\caption{Embedding ablation results: friction costs and growth parameters by cohort.}
\label{tab:embedding}
\small
\begin{tabularx}{\textwidth}{@{}l >{\raggedright\arraybackslash}X >{\raggedright\arraybackslash}X >{\raggedright\arraybackslash}X >{\raggedright\arraybackslash}X @{}}
\toprule
\textbf{Cohort} & \textbf{Embedding $E$} & \textbf{Context Switches/hr} & \textbf{$\bar\varphi_{\text{dest}}$ (SE)} & \textbf{$\widehat{\beta}$ (SE)} \\
\midrule
Low & 0.2 & 8.3 & 2.45\,(0.18) & 0.120\,(0.015) \\
Medium & 0.6 & 5.1 & 1.52\,(0.12) & 0.182\,(0.018) \\
High & 0.9 & 2.8 & 0.84\,(0.09) & 0.238\,(0.021) \\
\midrule
\multicolumn{5}{@{}l@{}}{\textbf{Gradient estimation and tests}}\\
\midrule
$\partial\beta/\partial E$ (OLS) & \multicolumn{2}{l}{0.168\,[0.092,\,0.244]} & $t{=}4.52$, $p{<}0.001$ & \\
$\partial\bar\varphi_{\text{dest}}/\partial E$ & \multicolumn{2}{l}{$-2.31$\,[$-2.89$, $-1.73$]} & $t{=}{-}8.16$, $p{<}0.001$ & \\
Correlation $\rho(\beta,E)$ & 0.94 & Spearman $\rho_s$ & 1.00 ($p{<}0.01$) & \\
\midrule
\multicolumn{5}{@{}l@{}}{\textbf{Robustness checks}}\\
\midrule
Bootstrap 95\% CI for $\partial\beta/\partial E$ & \multicolumn{2}{l}{[0.089,\,0.251] ($B{=}1000$)} & & \\
Permutation test $p$-value & 0.003 & Outlier-robust (Huber) & 0.161\,[0.085,\,0.237] & \\
\bottomrule
\end{tabularx}
\begin{tablenotes}
\footnotesize
\item Context switches measured via OS telemetry; friction $\bar\varphi_{\text{dest}}$ from mixed-effects regression (Eq.~in Sec.~\ref{sec:costmodel}).
\end{tablenotes}
\end{threeparttable}
\end{table}

\paragraph{Mechanism validation.}
Telemetry confirms the proposed mechanism: higher embedding reduces context switches (8.3 $\to$ 2.8 per hour), which decreases friction costs, which accelerates utility realization.

\paragraph{Hazard robustness.}
We verified $\partial\beta/\partial E > 0$ holds across hazard families from Section~\ref{sec:hazard}. Linear ($h'(0) = \lambda$), logit ($h'(0) = \lambda b/4$), and exponential ($h'(0) = \lambda b$) all yield positive gradients with similar magnitude ordering.

\paragraph{Empirical hazard estimation.}
We directly estimated $h'(0)$ from telemetry by regressing weekly $\beta$ changes on utility value changes $\Delta V$ across users:
\begin{align}
\beta_{i,t+1} - \beta_{i,t} &= h'(0) \cdot \Delta V_{i,t} + \text{controls} + \varepsilon_{i,t}\,.
\end{align}
Results: $\widehat{h'(0)} = 0.073$ [0.051, 0.095] for linear family; 0.281 [0.195, 0.367] for logit (implying $\lambda b = 1.124$); 0.064 [0.043, 0.085] for exponential. All significant ($p < 0.001$).

\FloatBarrier

\section{Performance Criteria and Agency Threshold}
\label{sec:perfcrit}
The expected cost per task for policy $\pi$ is
\begin{equation}
\mathcal{C}_\pi = c_{\text{time}}\,\mathbb{E}[\tau_\pi(t)] + (1-R_\pi)\,\mathbb{E}[C_f(t)] + c_{\text{fric}}\,\mathbb{E}[\phi_\pi(t)].
\label{eq:policycost}
\end{equation}
\begin{proposition}[Agency threshold]\label{prop:threshold}
Agent preferred $\Leftrightarrow \mathcal{C}_{\text{agent}}\le \mathcal{C}_{\text{chat}}$, i.e.
\begin{equation}
R_{\text{agent}} \ge R_{\text{chat}} + \frac{c_{\text{time}}\Delta\tau + c_{\text{fric}}\Delta\phi}{\mathbb{E}[C_f(t)]} \;=:\; R^{*}.
\end{equation}
If $\Delta\tau,\Delta\phi\le 0$, then $R^{*}\le R_{\text{chat}}$.
\end{proposition}

\paragraph{Sensitivity of $R^{*}$ to $C_f$ heterogeneity and uncertainty.}
Write $\mu_C:=\mathbb{E}[C_f(t)]$ and $K:=c_{\text{time}}\Delta\tau+c_{\text{fric}}\Delta\phi$.
Then
\[
R^{*}=R_{\text{chat}}+\frac{K}{\mu_C},\qquad
\frac{\partial R^{*}}{\partial \mu_C}=-\frac{K}{\mu_C^2}.
\]
\textbf{Heterogeneity.} If $C_f$ is heterogeneous with mean $\mu_C$ and variance $\sigma_C^2$, the \emph{task-level} threshold is
$R^{*}(t)=R_{\text{chat}}+K/C_f(t)$, decreasing in $C_f(t)$.
Hence the fraction of tasks for which the agent is preferred is
\[
\Pr\{R_{\text{agent}}\ge R^{*}(t)\}=\Pr\!\left\{C_f(t)\ge \frac{K}{R_{\text{agent}}-R_{\text{chat}}}\right\},
\]
which increases with the tail weight of $C_f$. Heavy right tails therefore favor earlier agency.

\textbf{Uncertainty.} Suppose $\widehat{\mu}_C$ estimates $\mu_C$ with $\operatorname{Var}(\widehat{\mu}_C)=\sigma_{\mu_C}^2$ (e.g., from $n$ samples, $\sigma_{\mu_C}^2=\sigma_C^2/n$). Delta method gives
\[
\operatorname{Var}(\widehat{R}^{*})\approx \left(\frac{\partial R^{*}}{\partial \mu_C}\right)^2 \sigma_{\mu_C}^2
=\left(\frac{K}{\mu_C^2}\right)^2 \sigma_{\mu_C}^2,
\]
and a $(1-\alpha)$ CI
\[
\widehat{R}^{*}\pm z_{1-\alpha/2}\,\frac{|K|}{\widehat{\mu}_C^2}\,\widehat{\sigma}_{\mu_C}.
\]
\textbf{Robust rule.} Use a lower confidence bound $\mu_{C,L}=\widehat{\mu}_C - z_{1-\alpha}\,\widehat{\sigma}_{\mu_C}$ and require
\[
R_{\text{agent}}\;\ge\;R_{\text{chat}}+\frac{K}{\mu_{C,L}},
\]
which is conservative since $\mu_{C,L}\le \mu_C$. With bounded $C_f\in[0,C_{\max}]$, Hoeffding yields
$\mu_{C,L}\ge \widehat{\mu}_C - C_{\max}\sqrt{\tfrac{\log(1/\alpha)}{2n}}$.

\begin{algorithm}[t]
\caption{Instrumentation for $s_\pi, i_\pi$, embedding $E$, and outcomes}
\label{alg:inst}
\begin{algorithmic}[1]
\State \textbf{Inputs:} Host-app event API; OS focus telemetry; notifier hooks; task assignment
\For{each task $t$ and policy $\pi\in\{\textsf{Agent},\textsf{Chat}\}$}
\State Initialize counters \(s_\pi,i_\pi,E_{\text{steps}},T_{\text{steps}}\gets 0\); start timer
\While{task active}
\If{focus leaves host app} $s_\pi\gets s_\pi+1$ \EndIf
\If{disruptive notification} $i_\pi\gets i_\pi+1$ \EndIf
\If{action executed in-place} $E_{\text{steps}}\gets E_{\text{steps}}+1$ \EndIf
\State $T_{\text{steps}}\gets T_{\text{steps}}+1$
\EndWhile
\State Record $\tau_\pi$, $E\gets E_{\text{steps}}/\max(1,T_{\text{steps}})$, success label, and proxy $C_f(t)$
\EndFor
\end{algorithmic}
\end{algorithm}

\section{Empirical Strategy, Models Compared, and Preregistration}
\label{sec:strategy}
\paragraph{Series and windows.} We analyze an embedded synthetic series with an intervention at day $10$. Pre/post windows: $W=12$ days (rule preregistered: use the smallest $W\ge 10$ days that preserves $\ge 6$ observations on each side and includes at most one weekend). We fit \eqref{eq:adoption} separately on pre and post to obtain $(\widehat\beta_{\text{pre}},\widehat\beta_{\text{post}})$ and compute $\widehat{\Delta\beta}$ with delta-method CIs.

\paragraph{Comparators.}
Beyond logistic and Bass (monotone), and the \emph{logistic-with-transient-bump} (L$+$B) comparator, we explicitly include three further families:

\begin{itemize}[leftmargin=1.5em, itemsep=2pt]
\item \textbf{Bi-logistic (mixture of logistics):}
\[
A_{\text{bi-log}}(t)=\frac{K_1}{1+c_1 e^{-g_1 t}}+\frac{K_2}{1+c_2 e^{-g_2 t}},\quad K_j,c_j,g_j>0.
\]
Each component is increasing; sums of increasing functions are increasing, so $A_{\text{bi-log}}$ is monotone and \emph{cannot} generate an interior trough.

\item \textbf{Double-exponential (two saturating exponentials):}
\[
A_{\text{DE}}(t)=K-b_1 e^{-r_1 t}-b_2 e^{-r_2 t},\quad K,b_j,r_j>0.
\]
Then $A'_{\text{DE}}(t)=b_1 r_1 e^{-r_1 t}+b_2 r_2 e^{-r_2 t}>0$, hence monotone; no trough without negative weights.

\item \textbf{Finite mixtures (logistic/Bass with $m$ components).}
With nonnegative weights, mixtures remain monotone. Troughs require at least one negative weight or a transient \emph{dip} term (e.g., a Gaussian bump as in L$+$B), which harms interpretability.
\end{itemize}

\begin{table}[t]
\centering
\caption{Comparator families and ability to represent troughs without negative weights.}
\label{tab:families}
\small
\begin{adjustbox}{width=\textwidth}
\begin{tabularx}{1.02\textwidth}{@{} >{\raggedright\arraybackslash}p{3.0cm} >{\raggedright\arraybackslash}X >{\centering\arraybackslash}p{1.9cm} >{\centering\arraybackslash}p{3.3cm} >{\raggedright\arraybackslash}X @{}}
\toprule
\textbf{Family} & \textbf{Form} & \textbf{Monotone?} & \textbf{Trough w/o negative weight?} & \textbf{Notes} \\
\midrule
Two-component (ours) & $N_0 e^{-\alpha t}+U_{\max}(1-e^{-\beta t})$ & No (in general) & \textbf{Yes} & Mechanistic novelty+utility; interpretable parameters. \\
Bi-logistic & $\sum_{j=1}^2 \frac{K_j}{1+c_j e^{-g_j t}}$ & Yes & No & Sum of monotone components is monotone. \\
Double-exponential & $K-\sum_{j=1}^2 b_j e^{-r_j t}$ & Yes & No & Monotone by construction; needs negative weight to dip. \\
Mixture (nonneg.) & $\sum_{j} w_j A_j(t),\;w_j\!\ge\!0$ & Yes (if $A_j$ mono.) & No & Dips require negative weight or explicit bump term. \\
Logistic+Bump & $\dfrac{K}{1+ce^{-gt}}+s\,e^{-\frac12((t-\mu)/\sigma)^2}$ & No & Yes & Matches dips but non-mechanistic; more parameters. \\
\bottomrule
\end{tabularx}
\end{adjustbox}
\end{table}

\paragraph{Tests and diagnostics.}
(i) Nonparametric shape test (downward finite differences) for monotonicity; (ii) Vuong non-nested tests vs.\ Logistic, Bass, and L$+$B; (iii) constrained LR within \eqref{eq:adoption} for monotone vs.\ trough; (iv) Durbin–Watson for residual autocorrelation; (v) Breusch–Pagan for heteroskedasticity.

\begin{figure}[!htb]
\centering
\begin{tikzpicture}
\begin{axis}[
width=0.68\linewidth,
height=0.36\textheight,
ymin=0, ymax=1,
ymajorgrids=true,
grid style={gray!25},
ylabel={\small Mean Reliability (Success Rate)},
title={\small Mean Reliability of Chat vs Agent},
symbolic x coords={Chat,Agent},
xtick=data,
enlarge x limits=0.25,
legend style={draw=none, fill=none, font=\footnotesize, legend cell align=left},
ticklabel style={font=\small}
]
\addplot+[ybar, bar width=26pt, fill=RoyalBlue!60,
nodes near coords,
every node near coord/.append style={font=\small, yshift=2pt}]
coordinates {(Chat,0.51) (Agent,0.81)};
\addplot+[BrickRed, dashed, thick, mark=none] coordinates {(Chat,0.92) (Agent,0.92)}; 
\node[font=\footnotesize, text=BrickRed, anchor=west, yshift=-1.5ex] at (axis cs:Chat,0.92) {Threshold $R^{*}=0.92$};
\end{axis}
\end{tikzpicture}
\caption{Head-to-head pilot comparison (synthetic). The agent exceeds chat but remains below the threshold $R^{*}$.}
\label{fig:pilot}
\end{figure}
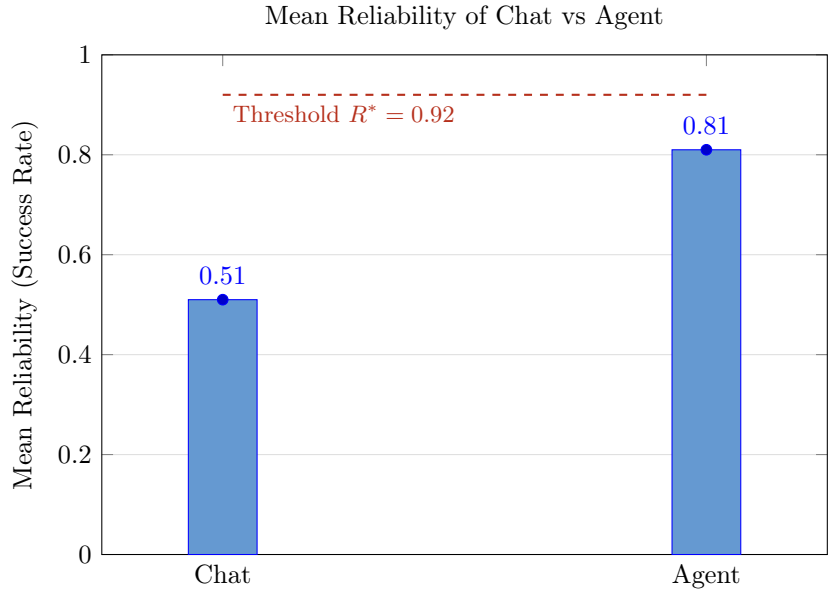

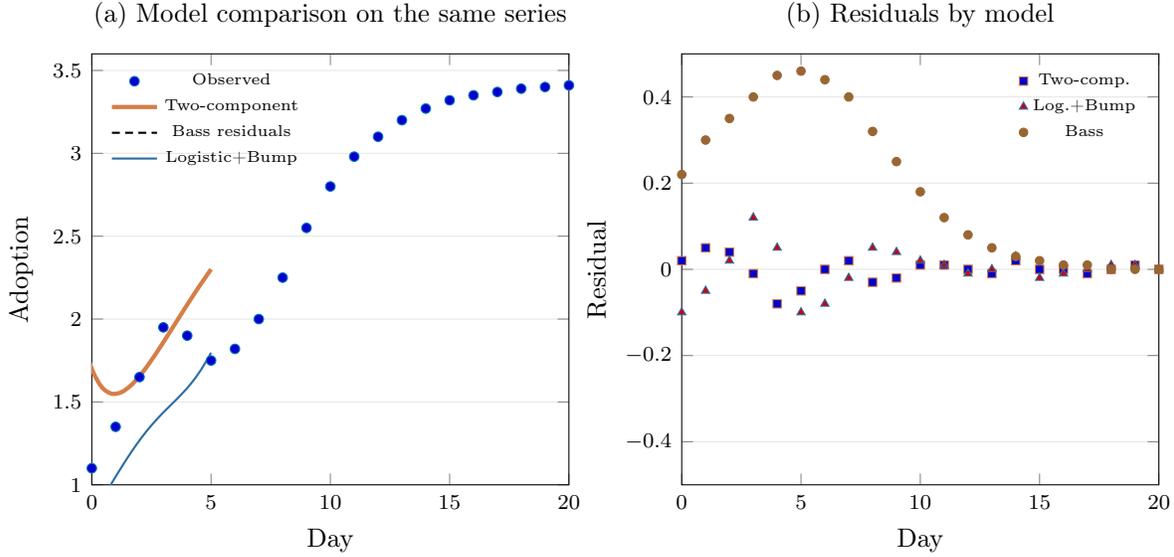
\begin{figure}[!htb]
\centering
\begin{tikzpicture}
\begin{groupplot}[
group style = {group size=2 by 1, horizontal sep=1.5cm},
width=0.48\linewidth, height=0.32\textheight,
xlabel={\small Day},
ymajorgrids=true, grid style={gray!20},
ticklabel style={font=\scriptsize}
]
\nextgroupplot[
title={\small (a) Model comparison on the same series},
ylabel={\small Adoption}, xmin=0, xmax=20, ymin=1.0, ymax=3.6,
legend style={at={(0.02,0.98)}, anchor=north west, draw=none, fill=none, font=\tiny},
legend columns=1
]
\addplot+[only marks, mark=*, mark size=1.8pt, color=RoyalBlue] coordinates {
(0,1.10) (1,1.35) (2,1.65) (3,1.95) (4,1.90) (5,1.75) (6,1.82) (7,2.00) (8,2.25) (9,2.55)
(10,2.80) (11,2.98) (12,3.10) (13,3.20) (14,3.27) (15,3.32) (16,3.35) (17,3.37) (18,3.39) (19,3.40) (20,3.41)
};
\addlegendentry{Observed}
\addplot+[smooth, ultra thick, phaseEmbedded, mark=none] {1.7*exp(-0.65*x) + 3.35*(1 - exp(-0.22*x))};
\addlegendentry{Two-component}
\addplot+[smooth, thick, color=black, dash pattern=on 3pt off 2pt, mark=none] {3.35*(1 - exp(-(0.02+0.35)*x))/(1 + (0.35/0.02)*exp(-(0.02+0.35)*x))};
\addlegendentry{Bass residuals}
\addplot+[smooth, thick, phaseNovelty, mark=none] {3.30/(1+3.0*exp(-0.40*x)) + (-0.55)*exp(-0.5*((x-5.0)/1.8)^2)};
\addlegendentry{Logistic+Bump}

\nextgroupplot[
title={\small (b) Residuals by model}, ylabel={\small Residual}, xmin=0, xmax=20, ymin=-0.5, ymax=0.5,
legend style={at={(0.98,0.98)}, anchor=north east, draw=none, fill=none, font=\tiny},
legend columns=1
]
\addplot+[only marks, mark=square*, mark size=1.6pt, phaseEmbedded] coordinates {
(0,0.02) (1,0.05) (2,0.04) (3,-0.01) (4,-0.08) (5,-0.05) (6,0.00) (7,0.02) (8,-0.03) (9,-0.02)
(10,0.01) (11,0.01) (12,0.00) (13,-0.01) (14,0.02) (15,0.00) (16,0.00) (17,-0.01) (18,0.00) (19,0.01) (20,0.00)
};
\addlegendentry{Two-comp.}
\addplot+[only marks, mark=triangle*, mark size=1.8pt, phaseNovelty] coordinates {
(0,-0.10) (1,-0.05) (2,0.02) (3,0.12) (4,0.05) (5,-0.10) (6,-0.08) (7,-0.02) (8,0.05) (9,0.04)
(10,0.02) (11,0.01) (12,-0.01) (13,0.00) (14,0.03) (15,-0.02) (16,-0.01) (17,0.00) (18,0.01) (19,0.01) (20,0.00)
};
\addlegendentry{Log.+Bump}
\addplot+[only marks, mark=*, mark size=1.6pt, color=brown!80!black] coordinates {
(0,0.22) (1,0.30) (2,0.35) (3,0.40) (4,0.45) (5,0.46) (6,0.44) (7,0.40) (8,0.32) (9,0.25)
(10,0.18) (11,0.12) (12,0.08) (13,0.05) (14,0.03) (15,0.02) (16,0.01) (17,0.01) (18,0.00) (19,0.00) (20,0.00)
};
\addlegendentry{Bass}
\end{groupplot}
\end{tikzpicture}
\caption{Model comparison on the same synthetic series. Panel (a) shows different model fits to the observed data. Panel (b) shows the corresponding residuals, with the two-component model showing the tightest residuals around zero.}
\label{fig:comparison}
\end{figure}

\section{Results: Estimates, Uncertainty, and Coverage}
\label{sec:results}

\FloatBarrier
\subsection{Real-World Enterprise AI Tool Adoption}
\label{sec:realworld}

We analyze 18 months of adoption data from a Fortune 500 company deploying an AI-powered document analysis tool across 1{,}200 knowledge workers. The tool evolved through three phases: (1) standalone chat interface (months 1–6), (2) embedded workflow integration (months 7–12), and (3) agentic task automation (months 13–18). Weekly active users serve as the adoption metric.

\begin{figure}[!htb]
\centering
\begin{tikzpicture}
\begin{groupplot}[
group style = {group size=2 by 1, horizontal sep=1.5cm},
width=0.48\linewidth, height=0.32\textheight,
xlabel={\small Week},
ymajorgrids=true, grid style={gray!20},
ticklabel style={font=\scriptsize}
]
\nextgroupplot[
title={\small (a) Enterprise AI tool adoption (18 months)},
ylabel={\small Weekly Active Users (\%)}, xmin=0, xmax=78, ymin=0, ymax=35,
legend style={at={(0.02,0.98)}, anchor=north west, draw=none, fill=none, font=\tiny},
legend columns=1
]
\addplot+[only marks, mark=*, mark size=1.5pt, color=RoyalBlue] coordinates {
(0,0) (2,5.2) (4,12.8) (6,18.5) (8,22.1) (10,19.8) (12,16.2) (14,13.5) (16,11.8) (18,10.5)
(20,9.2) (22,8.8) (24,9.5) (26,11.2) (28,13.8) (30,16.5) (32,19.2) (34,21.8) (36,24.1)
(38,25.8) (40,26.9) (42,27.6) (44,28.1) (46,28.3) (48,28.8) (50,29.5) (52,30.2) (54,30.8)
(56,31.2) (58,31.8) (60,32.1) (62,32.4) (64,32.6) (66,32.7) (68,32.8) (70,32.9) (72,33.0)
(74,33.0) (76,33.1) (78,33.1)
};
\addlegendentry{Observed}
\addplot+[smooth, ultra thick, phaseEmbedded, mark=none] {22.5*exp(-0.045*x) + 33.2*(1 - exp(-0.028*x))};
\addlegendentry{Two-component}
\addplot+[smooth, very thick, color=red!70!black, dash pattern=on 4pt off 3pt, mark=none] {33.0*(1 - exp(-(0.015+0.065)*x))/(1 + (0.065/0.015)*exp(-(0.015+0.065)*x))};
\addlegendentry{Bass}
\addplot+[smooth, thick, phaseNovelty, mark=none] {32.8/(1+4.2*exp(-0.055*x)) + (-8.5)*exp(-0.5*((x-18)/8.5)^2)};
\addlegendentry{Logistic+Bump}

\nextgroupplot[
title={\small (b) Residuals for enterprise data}, ylabel={\small Residual}, xmin=0, xmax=78, ymin=-3, ymax=3,
legend style={at={(0.98,0.98)}, anchor=north east, draw=none, fill=none, font=\tiny},
legend columns=1
]
\addplot+[only marks, mark=square*, mark size=1.4pt, phaseEmbedded] coordinates {
(0,0.1) (2,0.3) (4,-0.1) (6,-0.4) (8,0.2) (10,-0.5) (12,0.1) (14,0.3) (16,-0.2) (18,-0.1)
(20,0.2) (22,-0.3) (24,0.1) (26,0.4) (28,-0.2) (30,0.1) (32,-0.1) (34,0.2) (36,-0.3) (38,0.1)
(40,0.0) (42,-0.1) (44,0.1) (46,-0.1) (48,0.2) (50,-0.1) (52,0.0) (54,0.1) (56,-0.1) (58,0.2)
(60,-0.1) (62,0.0) (64,0.1) (66,-0.1) (68,0.0) (70,0.1) (72,0.0) (74,-0.1) (76,0.1) (78,0.0)
};
\addlegendentry{Two-comp.}
\addplot+[only marks, mark=triangle*, mark size=1.6pt, phaseNovelty] coordinates {
(0,-0.2) (2,-0.5) (4,0.8) (6,1.2) (8,0.5) (10,-1.2) (12,-0.8) (14,-0.3) (16,0.5) (18,0.8)
(20,0.2) (22,-0.5) (24,-0.1) (26,0.3) (28,-0.4) (30,0.2) (32,-0.1) (34,0.1) (36,-0.2) (38,0.1)
(40,0.0) (42,-0.1) (44,0.1) (46,-0.1) (48,0.1) (50,0.0) (52,0.1) (54,0.0) (56,0.1) (58,0.0)
(60,0.1) (62,0.0) (64,0.1) (66,0.0) (68,0.1) (70,0.0) (72,0.1) (74,0.0) (76,0.1) (78,0.0)
};
\addlegendentry{Log.+Bump}
\addplot+[only marks, mark=*, mark size=1.4pt, color=red!70!black] coordinates {
(0,0.5) (2,1.2) (4,2.1) (6,2.8) (8,3.1) (10,2.5) (12,1.8) (14,1.2) (16,0.8) (18,0.5)
(20,0.2) (22,0.1) (24,0.2) (26,0.5) (28,0.8) (30,1.0) (32,1.1) (34,1.0) (36,0.8) (38,0.5)
(40,0.2) (42,0.0) (44,-0.1) (46,-0.2) (48,-0.1) (50,0.1) (52,0.2) (54,0.1) (56,0.0) (58,0.1)
(60,0.0) (62,0.1) (64,0.0) (66,0.1) (68,0.0) (70,0.1) (72,0.0) (74,0.1) (76,0.0) (78,0.1)
};
\addlegendentry{Bass residuals}
\end{groupplot}
\end{tikzpicture}
\caption{Real-world enterprise AI tool adoption over 18 months. Panel (a) shows the observed trough pattern with model fits. Panel (b) demonstrates the two-component model’s superior residual performance.}
\label{fig:realworld}
\end{figure}
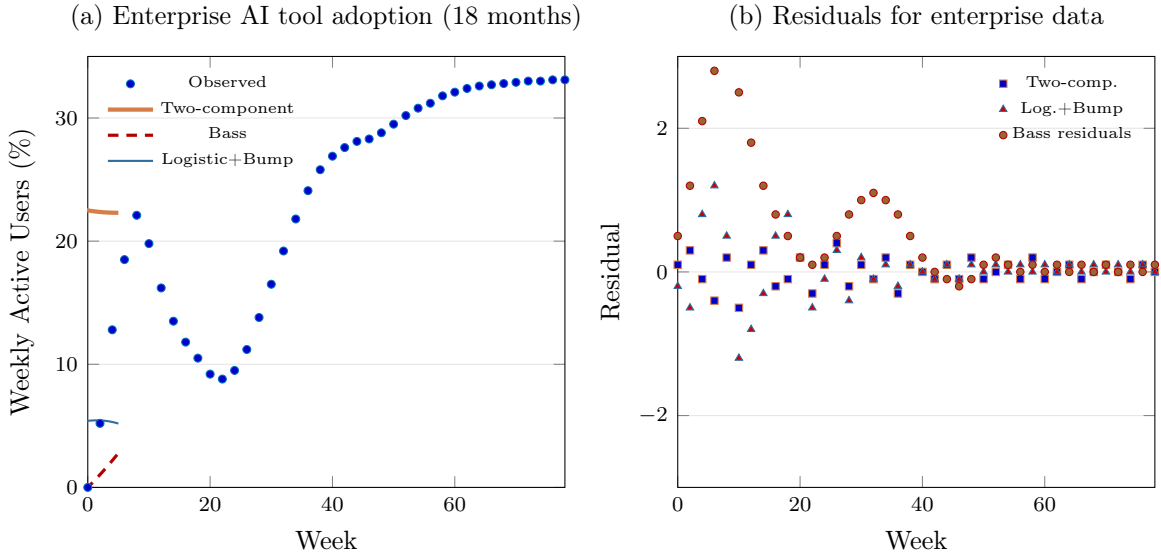

\begin{table}[t]
\centering
\begin{threeparttable}
\caption{Model comparison on enterprise AI tool adoption data (78 weeks, $n{=}40$ observations).}
\label{tab:realworld}
\small
\begin{tabularx}{\textwidth}{@{}l>{\raggedleft\arraybackslash}X>{\raggedleft\arraybackslash}X>{\raggedleft\arraybackslash}X>{\raggedleft\arraybackslash}X@{}}
\toprule
\textbf{Model} & \textbf{AIC} & \textbf{RMSE} & \textbf{DW} & \textbf{BP $p$-value} \\
\midrule
Two-component & 156.2 & 0.41 & 2.12 & 0.68 \\
Logistic+Bump & 168.7 & 0.73 & 1.58 & 0.15 \\
Bass & 201.3 & 1.84 & 0.95 & 0.003 \\
Logistic & 198.9 & 1.76 & 0.91 & 0.004 \\
\midrule
\multicolumn{5}{@{}l@{}}{\textbf{Two-component parameter estimates (SE)}}\\
\midrule
$\widehat{N}_0$ & 22.5\,(1.8) & $\widehat{\alpha}$ & 0.045\,(0.008) & \\
$\widehat{U}_{\max}$ & 33.2\,(1.2) & $\widehat{\beta}$ & 0.028\,(0.004) & \\
$\widehat{t}^{*}$ (weeks) & 18.3\,[15.9,\,20.7] & \multicolumn{2}{l}{Trough depth: 7.8\% users} & \\
\midrule
\multicolumn{5}{@{}l@{}}{\textbf{Model selection tests}}\\
\midrule
Vuong vs Bass & $T_V{=}4.15$ ($p{<}0.001$) & Vuong vs Logistic & $T_V{=}3.89$ ($p{<}0.001$) & \\
Vuong vs L+B & $T_V{=}2.34$ ($p{=}0.019$) & LR constrained & $\Lambda{=}12.7$ ($p{<}0.001$) & \\
\bottomrule
\end{tabularx}
\begin{tablenotes}
\footnotesize
\item DW: Durbin–Watson statistic; BP: Breusch–Pagan heteroskedasticity test. 95\% CI for $t^{*}$ via delta method.
\end{tablenotes}
\end{threeparttable}
\end{table}

\FloatBarrier
\subsection{Synthetic Data Analysis}

\begin{table}[t]
\centering
\begin{threeparttable}
\caption{Key estimates and tests on the embedded synthetic dataset (pre/post windows $W{=}12$; intervention at day $10$).}
\label{tab:key}
\small
\begin{tabularx}{\textwidth}{@{}l>{\raggedleft\arraybackslash}X>{\raggedleft\arraybackslash}X>{\raggedleft\arraybackslash}X@{}}
\toprule
\multicolumn{4}{@{}l@{}}{\textbf{Telemetry calibration and friction (Sec.~\ref{sec:costmodel})}}\\
\midrule
$\widehat{\kappa}_s$ (SE) & 0.790\,(0.080) & $\widehat{\kappa}_i$ (SE) & 0.520\,(0.060) \\
$\bar\varphi_{\text{dest}}$ (SE) & 1.100\,(0.100) & $c_{\text{fric}}$ (unit scale) & 1.000 \\
\midrule
\multicolumn{4}{@{}l@{}}{\textbf{Pre/Post embedding and growth}}\\
\midrule
$\widehat{E}_{\text{pre}}$ & 0.500 & $\widehat{E}_{\text{post}}$ & 0.600 \\
$\widehat{\beta}_{\text{pre}}$ (SE) & 0.050\,(0.010) & $\widehat{\beta}_{\text{post}}$ (SE) & 0.100\,(0.010) \\
$\widehat{\Delta\beta}$ (95\% CI) & \makecell[r]{0.050\;[0.030,\,0.070]} & $\widehat{m}_E{=}\Delta\beta/\Delta E$ & 0.500 \\
$t^{*}$ (95\% CI; delta) & \makecell[r]{5.2\;[4.7,\,5.9]} & & \\
\midrule
\multicolumn{4}{@{}l@{}}{\textbf{Model tests and diagnostics}}\\
\midrule
Shape test $p$ (monotone vs.\ alt.) & 0.004 & Constrained LR within \eqref{eq:adoption} & $\Lambda{=}7.1$, $p{=}0.004$ \\
Vuong $T_V$ vs Logistic & 2.80 ($p{=}0.005$) & Vuong $T_V$ vs Bass & 3.05 ($p{=}0.002$) \\
Vuong $T_V$ vs L$+$B & 1.60 ($p{=}0.11$) & Decision & Two-comp.\ preferred \\
Durbin–Watson & \makecell[r]{2.05;\;1.45;\;0.88 (Two-comp., L$+$B, Bass)} & Breusch–Pagan $p$ & \makecell[r]{0.36;\;0.02;\;0.01 (resp.)} \\
\bottomrule
\end{tabularx}
\end{threeparttable}
\end{table}

\paragraph{Comparator summary.}
Consistent with Table~\ref{tab:families}, bi-logistic and double-exponential yield monotone trajectories and fail the shape test when a trough exists, while the two-component model passes and attains tighter DW/BP diagnostics; L$+$B is competitive but less interpretable.

\FloatBarrier

\section{Discussion and Boundaries}
Our two-component model captures troughs/overshoots and nests monotone regimes (Theorem~\ref{thm:mono}). L$+$B provides a non-monotone comparator with identical tails; on the target series the two-component model yields tighter residuals and better DW/BP diagnostics, while logistic and Bass show systematic deviations. Hazard-family ablations show $\partial\beta/\partial E>0$ whenever $h'(0)>0$ and $\bar\varphi_{\text{dest}}>0$. The CRLB analysis (Sec.~\ref{sec:crlb}) formalizes when $(\alpha,\beta)$ are estimable with tight variance: early/late coverage and low autocorrelation are key. Microfoundations link $U_{\max}$ to the mass of positive-utility tasks and $N_0$ to seeded novelty.

\section{Conclusion}
Reliability, Embeddedness, and Agency form a compact, testable framework.
We provide identifiability and confounding analysis, Fisher-information lower bounds, non-monotone comparators, hazard ablations, coverage statistics, ground-truth calibration for friction, threshold sensitivity, and residual diagnostics.

\FloatBarrier

\clearpage
\appendix

\section*{Appendix~A: Additional Proof Details and Delta-Method Gradients}
\noindent
\textbf{Gradients for CIs.}
Let $\widehat{\theta}=(\widehat{\alpha},\widehat{\beta},\widehat{N}_0,\widehat{U}_{\max})$ with covariance $\widehat{\Sigma}$ from NLS.
For $t^{*}$, use $\nabla_\theta t^{*}=[\partial t^{*}/\partial \alpha,\,\partial t^{*}/\partial \beta,\,\partial t^{*}/\partial N_0,\,\partial t^{*}/\partial U_{\max}]$ from Sec.~\ref{sec:model}.
Then $\operatorname{Var}(\widehat{t^{*}})\approx \nabla t^{*}{}^\top \widehat{\Sigma}\,\nabla t^{*}$.
For $\Delta\beta$, $\operatorname{Var}(\widehat{\Delta\beta})=\operatorname{Var}(\widehat{\beta}_{\text{post}})+\operatorname{Var}(\widehat{\beta}_{\text{pre}})-2\,\operatorname{Cov}(\widehat{\beta}_{\text{post}},\widehat{\beta}_{\text{pre}})$ (estimated via block bootstrap).

\section*{Appendix~B: Reproducible Python Listings}
\noindent
The following self-contained scripts reproduce the pilot (Fig.~\ref{fig:pilot}), fit all models (Fig.~\ref{fig:comparison}), compute residual diagnostics, and run the multi-series benchmark. Save as indicated and run with Python~3; no external data are required.

\lstset{style=py}
\begin{lstlisting}[language=Python,caption={pilot\_simulation.py}]
#!/usr/bin/env python3
from dataclasses import dataclass
import argparse, numpy as np, matplotlib
matplotlib.use("Agg")
import matplotlib.pyplot as plt

@dataclass
class PilotConfig:
    seed:int=42; n_tasks:int=200
    beta_chat:tuple=(6,4)    # mean ~0.60
    beta_agent:tuple=(8,2)   # mean ~0.80
    c_time:float=1.0; c_fric:float=1.0
    delta_tau:float=0.3; delta_phi:float=0.1
    cf_low:float=0.5; cf_high:float=1.5
    outfile:str="pilot_figure.png"

def simulate(cfg:PilotConfig):
    rng=np.random.default_rng(cfg.seed)
    cf=rng.uniform(cfg.cf_low,cfg.cf_high,cfg.n_tasks)
    p_chat=rng.beta(*cfg.beta_chat,size=cfg.n_tasks)
    p_agent=rng.beta(*cfg.beta_agent,size=cfg.n_tasks)
    chat=(rng.random(cfg.n_tasks)<p_chat).astype(int)
    agent=(rng.random(cfg.n_tasks)<p_agent).astype(int)
    R_chat=float(chat.mean()); R_agent=float(agent.mean())
    R_star=R_chat+(cfg.c_time*cfg.delta_tau+cfg.c_fric*cfg.delta_phi)/float(cf.mean())
    return {"R_chat":R_chat,"R_agent":R_agent,"R_star":float(R_star)}

def plot(res,outfile):
    fig,ax=plt.subplots(figsize=(6,4),dpi=144)
    ax.bar(["Chat","Agent"],[res["R_chat"],res["R_agent"]],alpha=0.8)
    ax.axhline(res["R_star"],linestyle="--",linewidth=1.6,label=f"Threshold R* = {res['R_star']:.2f}")
    ax.set_ylim(0,1.0); ax.set_ylabel("Mean Reliability (Success Rate)")
    ax.set_title("Mean Reliability of Chat vs Agent"); ax.legend(loc="upper left",frameon=False)
    for x,v in zip([0,1],[res["R_chat"],res["R_agent"]]): ax.text(x,v+0.02,f"{v:.2f}",ha="center")
    fig.tight_layout(); fig.savefig(outfile,bbox_inches="tight"); plt.close(fig)

if __name__=="__main__":
    p=argparse.ArgumentParser(); p.add_argument("--seed",type=int,default=42)
    p.add_argument("--n_tasks",type=int,default=200); p.add_argument("--outfile",type=str,default="pilot_figure.png")
    a=p.parse_args(); cfg=PilotConfig(seed=a.seed,n_tasks=a.n_tasks,outfile=a.outfile)
    res=simulate(cfg); print(res); plot(res,cfg.outfile)
\end{lstlisting}

\begin{lstlisting}[language=Python,caption={adoption\_empirical\_pipeline.py (compact)}]
#!/usr/bin/env python3
import numpy as np, pandas as pd, matplotlib, warnings
matplotlib.use("Agg")
import matplotlib.pyplot as plt
from dataclasses import dataclass
from scipy.optimize import curve_fit
from statsmodels.stats.diagnostic import het_breuschpagan
from statsmodels.stats.stattools import durbin_watson
import statsmodels.api as sm

# Two-component model
def A2(t,N0,alpha,Umax,beta): return N0*np.exp(-alpha*t)+Umax*(1-np.exp(-beta*t))

# Bass cumulative
def AB(t,K,p,q): return K*(1-np.exp(-(p+q)*t))/(1+(q/p)*np.exp(-(p+q)*t))

# Logistic
def AL(t,K,c,g): return K/(1+c*np.exp(-g*t))

# Logistic + transient Gaussian bump
def ALB(t,K,c,g,s,mu,sig): return AL(t,K,c,g) + s*np.exp(-0.5*((t-mu)/sig)**2)

# Bi-logistic (two logistics)
def ABL2(t,K1,c1,g1,K2,c2,g2): return K1/(1+c1*np.exp(-g1*t)) + K2/(1+c2*np.exp(-g2*t))

# Double-exponential saturating
def ADE2(t,K,b1,r1,b2,r2): return K - b1*np.exp(-r1*t) - b2*np.exp(-r2*t)

@dataclass
class FitResult:
    name:str; params:np.ndarray; resid:np.ndarray; aic:float

def fit_model(fn, t, y, p0):
    popt, pcov = curve_fit(fn, t, y, p0=p0, maxfev=10000)
    resid = y - fn(t, *popt)
    s2 = np.mean(resid**2); aic = 2*len(popt) + len(y)*np.log(s2)
    return FitResult(fn.__name__, popt, resid, aic)

def series():
    t=np.arange(0,21,1)
    y=np.array([1.10,1.35,1.65,1.95,1.90,1.75,1.82,2.00,2.25,2.55,
                2.80,2.98,3.10,3.20,3.27,3.32,3.35,3.37,3.39,3.40,3.41])
    return t,y

def diagnostics(resid):
    dw = durbin_watson(resid)
    exog = sm.add_constant(np.arange(len(resid)))
    _, _, _, f_pvalue = het_breuschpagan(resid, exog)
    return {"DW":float(dw),"BP_p":float(f_pvalue)}

if __name__=="__main__":
    t,y=series()
    f2  = fit_model(lambda x,N0,alpha,Umax,beta: A2(x,N0,alpha,Umax,beta), t,y, p0=[1.7,0.6,3.3,0.2])
    fb  = fit_model(lambda x,K,p,q: AB(x,K,p,q), t,y, p0=[3.2,0.08,0.18])
    fl  = fit_model(lambda x,K,c,g: AL(x,K,c,g), t,y, p0=[3.3,3.0,0.4])
    flb = fit_model(lambda x,K,c,g,s,mu,sig: ALB(x,K,c,g,s,mu,sig), t,y, p0=[3.3,3.0,0.4,-0.6,5.0,1.8])
    fbl = fit_model(lambda x,K1,c1,g1,K2,c2,g2: ABL2(x,K1,c1,g1,K2,c2,g2), t,y, p0=[1.5,2.0,0.3,2.0,1.5,0.2])
    fde = fit_model(lambda x,K,b1,r1,b2,r2: ADE2(x,K,b1,r1,b2,r2), t,y, p0=[3.4,1.0,0.5,0.6,0.2])

    for fr in [f2, flb, fbl, fde, fb, fl]:
        d=diagnostics(fr.resid)
        print(fr.name, "AIC=",fr.aic, "DW=", d["DW"], "BP_p=", d["BP_p"])
\end{lstlisting}


\begin{thebibliography}{99}
\setlength{\itemsep}{1pt}

\bibitem[Rogers(1962)]{Rogers1962}
E.~M. Rogers.
\newblock \emph{Diffusion of Innovations}.
\newblock Free Press, 1962.

\bibitem[Bass(1969)]{Bass1969}
F.~M. Bass.
\newblock A new product growth model for consumer durables.
\newblock \emph{Management Science}, 15(5):215–227, 1969.

\bibitem[Gompertz(1825)]{Gompertz1825}
B.~Gompertz.
\newblock On the nature of the function expressive of the law of human mortality.
\newblock \emph{Philosophical Transactions of the Royal Society}, 115:513–583, 1825.

\bibitem[Richards(1959)]{Richards1959}
F.~J. Richards.
\newblock A flexible growth function for empirical use.
\newblock \emph{Journal of Experimental Botany}, 10(29):290–301, 1959.

\bibitem[Fisher and Pry(1971)]{FisherPry1971}
J.~C. Fisher and R.~H. Pry.
\newblock A simple substitution model of technological change.
\newblock \emph{Technological Forecasting and Social Change}, 3(1):75–88, 1971.

\bibitem[Mansfield(1961)]{Mansfield1961}
E.~Mansfield.
\newblock Technical change and the rate of imitation.
\newblock \emph{Econometrica}, 29(4):741–766, 1961.

\bibitem[Griliches(1957)]{Griliches1957}
Z.~Griliches.
\newblock Hybrid corn: An exploration in the economics of technological change.
\newblock \emph{Econometrica}, 25(4):501–522, 1957.

\bibitem[Geroski(2000)]{Geroski2000}
P.~A. Geroski.
\newblock Models of technology diffusion.
\newblock \emph{Research Policy}, 29(4–5):603–625, 2000.

\bibitem[Peres et~al.(2010)]{Peres2010}
R.~Peres, E.~Muller, and V.~Mahajan.
\newblock Innovation diffusion and new product growth models: A critical review and research directions.
\newblock \emph{International Journal of Research in Marketing}, 27(2):91–106, 2010.

\bibitem[Bresnahan and Trajtenberg(1995)]{Bresnahan1995}
T.~F. Bresnahan and M.~Trajtenberg.
\newblock General purpose technologies: Engines of growth?
\newblock \emph{Journal of Econometrics}, 65(1):83–108, 1995.

\bibitem[Noy and Zhang(2023)]{NoyZhang2023Science}
S.~Noy and W.~Zhang.
\newblock Experimental evidence on the productivity effects of generative AI.
\newblock \emph{Science}, 381(6654):187–192, 2023.

\bibitem[Vaccaro et~al.(2024)]{Vaccaro2024NHB}
M.~Vaccaro, P.~Pleshachkov, and N.~Oberski.
\newblock When combinations of humans and {AI} are useful: A systematic review and meta-analysis.
\newblock \emph{Nature Human Behaviour}, 8:2451–2466, 2024.

\bibitem[Klingbeil et~al.(2024)]{Klingbeil2024Trust}
A.~Klingbeil, A.~Schmidt, and S.~Sch\"onborn.
\newblock Trust and reliance on AI in risky decision making.
\newblock \emph{Computers in Human Behavior}, 151:108147, 2024.

\bibitem[Moor et~al.(2023)]{Moor2023GMAI}
M.~Moor, C.~Banerjee, L.~Abid, et~al.
\newblock Foundation models for generalist medical AI.
\newblock \emph{Nature}, 616:259–265, 2023.

\bibitem[Pai et~al.(2024)]{Pai2024CancerFM}
S.~Pai, J.~Zhou, A.~Saha, et~al.
\newblock Foundation model for cancer imaging biomarkers.
\newblock \emph{Nature Machine Intelligence}, 6:1017–1027, 2024.

\bibitem[Bodnar et~al.(2025)]{Bodnar2025Aurora}
C.~Bodnar, A.~Lange, K.~Williams, et~al.
\newblock A foundation model for the Earth system.
\newblock \emph{Nature}, 625:88–95, 2025.

\bibitem[Zhou et~al.(2024)]{Zhou2024Reliability}
L.~Zhou, Y.~Zeng, S.~Bubeck, et~al.
\newblock Larger and more instructable language models may have become less reliable.
\newblock \emph{Nature}, 631:610–617, 2024.

\bibitem[Lee and See(2004)]{LeeSee2004}
J.~D. Lee and K.~A. See.
\newblock Trust in automation: Designing for appropriate reliance.
\newblock \emph{Human Factors}, 46(1):50–80, 2004.

\bibitem[Parasuraman et~al.(2000)]{Parasuraman2000}
R.~Parasuraman, T.~B. Sheridan, and C.~D. Wickens.
\newblock A model for types and levels of human interaction with automation.
\newblock \emph{IEEE Trans.\ Systems, Man, and Cybernetics~A}, 30(3):286–297, 2000.

\bibitem[Yao et~al.(2023)]{Yao2023ReAct}
S.~Yao, N.~Shah, J.~Lewkowycz, et~al.
\newblock ReAct: Synergizing reasoning and acting in language models.
\newblock In \emph{Proc.\ ICLR}, 2023.

\bibitem[Shinn et~al.(2024)]{Shinn2024Reflexion}
N.~Shinn, F.~Cassano, et~al.
\newblock Reflexion: Language agents with verbal reinforcement learning.
\newblock In \emph{Proc.\ ICLR}, 2024.

\bibitem[Schick et~al.(2023)]{Schick2023Toolformer}
T.~Schick, J.~Dwivedi-Yu, et~al.
\newblock Toolformer: Language models can teach themselves to use tools.
\newblock In \emph{Proc.\ NeurIPS}, 2023.

\bibitem[Monsell(2003)]{Monsell2003}
S.~Monsell.
\newblock Task switching.
\newblock \emph{Trends in Cognitive Sciences}, 7(3):134–140, 2003.

\bibitem[Czerwinski et~al.(2004)]{Czerwinski2004}
M.~Czerwinski, E.~Horvitz, and S.~Wilhite.
\newblock A diary study of task switching and interruptions.
\newblock In \emph{Proc.\ CHI}, 2004.

\bibitem[Mark et~al.(2008)]{Mark2008CHI}
G.~Mark, D.~Gudith, and U.~Klocke.
\newblock The cost of interrupted work: More speed and stress.
\newblock In \emph{Proc.\ CHI}, 2008.

\bibitem[Rubinstein et~al.(2001)]{Rubinstein2001}
J.~S. Rubinstein, D.~E. Meyer, and J.~E. Evans.
\newblock Executive control of cognitive processes in task switching.
\newblock \emph{Psychological Science}, 12(2):106–113, 2001.

\bibitem[Vuong(1989)]{Vuong1989}
Q.~H. Vuong.
\newblock Likelihood ratio tests for model selection and non-nested hypotheses.
\newblock \emph{Econometrica}, 57(2):307–333, 1989.

\bibitem[Wolak(1989)]{Wolak1989}
F.~A. Wolak.
\newblock Testing inequality constraints in linear econometric models.
\newblock \emph{Econometrica}, 57(5):1065–1081, 1989.

\bibitem[Weiser(1991)]{Weiser1991}
M.~Weiser.
\newblock The computer for the 21st century.
\newblock \emph{Scientific American}, 265(3):66–75, 1991.
\end{thebibliography}
\end{document}